\documentclass[runningheads]{llncs}
\usepackage{graphicx,multirow}
\usepackage{textcomp}
\begin{document}
\title{FaceHop: A Light-Weight Low-Resolution Face Gender Classification Method}
%
%
\author{Mozhdeh Rouhsedaghat\inst{1},
Yifan Wang\inst{1}, Xiou Ge\inst{1},
Shuowen Hu\inst{2}, Suya You\inst{2},
and C.-C. Jay Kuo\inst{1}}
\authorrunning{F. Author et al.}
%
\institute{University of Southern California, Los Angeles, California, USA
\email{\{rouhseda,wang608,xiouge,jckuo\}@usc.edu}\\ \and
Army Research Laboratory, Adelphi, Maryland, USA\\
\email{\{shuowen.hu.civ,suya.you.civ\}@mail.mil}}
\maketitle              
\begin{abstract}
A light-weight low-resolution face gender classification method, called
FaceHop, is proposed in this research. We have witnessed rapid
progress in face gender classification accuracy due to the adoption of
deep learning (DL) technology.  Yet, DL-based systems are not suitable
for resource-constrained environments with limited networking and
computing.  FaceHop offers an interpretable non-parametric machine
learning solution. It has desired characteristics such as a small model
size, a small training data amount, low training complexity, and low-resolution input images. FaceHop is developed with the successive
subspace learning (SSL) principle and built upon the foundation of
PixelHop++.  The effectiveness of the FaceHop method is demonstrated by
experiments. For gray-scale face images of resolution $32 \times 32$ in
the LFW and the CMU Multi-PIE datasets, FaceHop achieves correct gender
classification rates of 94.63\% and 95.12\% with model sizes of 16.9K
and 17.6K parameters, respectively. It outperforms LeNet-5 in
classification accuracy while LeNet-5 has a model size of 75.8K
parameters. 

\end{abstract}

\section{Introduction}\label{sec:introduction}

Face attributes classification is an important topic in biometrics. The
ancillary information of faces such as gender, age, and ethnicity is
referred to as soft biometrics in forensics~\cite{Jain2004,riccio2012ega,Fierrez2018}.  The face gender classification problem has
been extensively studied for more than two decades. Before the
resurgence of deep neural networks (DNNs) around 7-8 years ago, the
problem was treated using the standard pattern recognition paradigm.  It
consists of two cascaded modules: 1) unsupervised feature extraction and
2) supervised classification via common machine learning tools such as
support vector machine (SVM) and random forest (RF) classifiers. 

We have seen fast progress on this topic due to the application of
deep learning (DL) technology in recent years. Generally speaking,
cloud-based face verification, recognition, and attributes classification
technologies have become mature, and they have been used in many real
world biometric systems.  Convolution neural networks (CNNs) offer high-performance accuracy. Yet, they rely on large learning models consisting
of several hundreds of thousands or even millions of model parameters.
The superior performance is contributed by factors such as higher input
image resolutions, more and more training images, and abundant
computational/memory resources. 

Edge/mobile computing in a resource-constrained environment cannot meet the above-mentioned conditions.  The technology of our interest finds
applications in rescue missions and/or field operational settings in
remote locations.  The accompanying face inference tasks are expected to
execute inside a poor computing and communication infrastructure.  It is
essential to have a smaller learning model size, lower training and
inference complexity, and lower input image resolution. The last
requirement arises from the need to image individuals at farther
standoff distances, which results in faces with fewer pixels. 

In this work, we propose a new interpretable non-parametric machine
learning solution called the FaceHop method. FaceHop has quite a few
desired characteristics, including a small model size, a small training
data amount, low training complexity, and low-resolution input images.
FaceHop follows the traditional pattern recognition paradigm that
decouples the feature extraction module from the decision module.
However, FaceHop automatically extracts statistical features instead of
handcrafted features.  It is developed with the successive subspace
learning (SSL) principle \cite{kuo2016understanding,kuo2017cnn,kuo2019interpretable} and built upon the foundation of 
the PixelHop++ system \cite{chen2020pixelhop++}. The effectiveness of the FaceHop method is demonstrated by experiments
on two benchmarking datasets.  For gray-scale face images of resolution
$32 \times 32$ obtained from the LFW and the CMU Multi-PIE datasets,
FaceHop achieves gender classification accuracy of 94.63\% and
95.12\% with model sizes of 16.9K and 17.6K parameters, respectively.
FaceHop outperforms LeNet-5 while the LeNet-5 model is significantly larger and contains 75.8K parameters. 

There are three main contributions of this work.  First, it offers a
practical solution to the challenging face biometrics problem in a
resource-constrained environment. Second, it is the first effort that
applies SSL to face gender classification and demonstrates its superior
performance. Third, FaceHop is fully interpretable, non-parametric, and
non-DL-based. It offers a brand new path for research and development in
biometrics. 

The rest of this paper is organized as follows. Related work is reviewed
in Sec. \ref{sec:review}. The FaceHop method is presented in Sec.
\ref{sec:method}. Experimental set-up and results are detailed in Sec.
\ref{sec:experiments}. Finally, concluding remarks and future extensions
are given in Sec.  \ref{sec:conclusion}. 

\section{Related Work}\label{sec:review}


\subsection{Face Attributes Classification}\label{subsec:gender}

We can classify face attributes classification research into two
categories: non-DL-based and DL-based. DL-based solutions construct an
end-to-end parametric model (i.e. a network), define a cost function,
and train the network to minimize the cost function with labeled face
gender images.  The contribution typically arises from a novel network
design.  Non-DL-based solutions follow the pattern recognition paradigm
and their contributions lie in using different classifiers or extracting new features for better
performance. 

{\bf Non-DL-based Solutions.} Researchers have studied different classifiers for gender classification. Gutta {\em et al.}~\cite{Phillips1998}
proposed a face-based gender and ethnic classification method using the
ensemble of Radial Basis Functions (RBF) and Decision Trees (DT).
SVM~\cite{moghaddam2002learning} and AdaBoost~\cite{baluja2007boosting} have been studied for face gender classification. 
Different feature extraction
techniques were experimented to improve classification accuracy. A
Gabor-kernel partial-least squares discrimination (GKPLSD) method for
more effective feature extraction was proposed by \v{S}truc {\em et
al.}~\cite{Vitomir2009}. Other handcrafted features were developed for
face gender classification based on the local directional patterns (LDP)
\cite{jabid2010gender} and shape from shading \cite{wu2010facial}.  Cao
{\em et al.}~\cite{Tan2015} combined Multi-order Local Binary Patterns
(MOLBP) with Localized Multi-Boost Learning (LMBL) for gender
classification. 

Recent research has focused more on large-scale face image datasets.  Li
{\em et al.}~\cite{Ruiping2015} proposed a novel binary code learning
method for large-scale face image retrieval and facial attribute
prediction.  Jia {\em et al.}~\cite{jia2015learning} collected a large
dataset of 4 million weakly labeled face in the wild (4MWLFW). They
trained the C-Pegasos classifier with Multiscale Local Binary Pattern
(LBP) features using the 4MWLFW dataset and achieved the highest test
accuracy on the LFW dataset for Non-DL-based methods up to now. Fusion
of different feature descriptors and region of interests (ROI) were
examined by Castrill{\'o}n-Santana {\em et al.}
\cite{castrillon2017descriptors}. 

The mentioned methods either have a weak performance as a result of failing to extract strong features from face images or have a large model size.

{\bf DL-based Solutions.} With the rapid advancement of the DL
technology, DL-based methods become increasingly popular and achieve
unprecedented accuracy in face biometrics \cite{lee2019facial}. 
Levi {\em et al.}~\cite{levi2015age} proposed a model to estimate age and gender using a small training data. Duan{\em et al.}~\cite{duan2018hybrid} introduced a hybrid CNN-ELM structure for age and gender classification which uses CNN for feature extraction from face images and ELM for classifying the features.
Taherkhani {\em et al.}~\cite{Taherkhani_2018_CVPR_Workshops} proposed a
deep framework which predicts facial attributes and leveraged it as a
soft modality to improve face identification performance. Han {\em et al.} \cite{Jain2018}investigated the heterogeneous face attribute estimation problem with a deep multi-task learning approach.  Ranjan {\em et al.} \cite{Patel2019}
proposed a multi-task learning framework for joint face detection,
landmark localization, pose estimation, and gender recognition.  
Antipov {\em et al.}~\cite{antipov2017effective} investigated the
relative importance of various regions of human faces for gender and age
classification by blurring different parts of the faces and observing
the loss in performance. ResNet50~\cite{antipov2017effective},
AlexNet~\cite{cheng2019exploiting}, and VGG16~\cite{lee2019facial} were
applied to gender classification of the LFW dataset, and decent
performance was observed. However, these models have very large
model sizes. Considerable amounts of computation and storage resources
are required to implement these solutions. 

{\bf Light-Weight CNNs.} Light-weight networks are significantly smaller
in size than regular networks while achieving comparable performance.
They find applications in mobile/edge computing.  One recent development
is the SqueezeNet \cite{iandola2016squeezenet} which achieves comparable
accuracy with the AlexNet \cite{krizhevsky2012imagenet} but uses
50x fewer parameters.  It contains 4.8M model parameters.  In the area of face recognition, Wu {\em et al.}
\cite{wu2018light} proposed a light CNN architecture that learns a
compact embedding on a large-scale face dataset with massive noisy
labels. Although the mentioned models are relatively small, they still require a large amount of training data.

\subsection{Successive Subspace Learning (SSL)}\label{subsec:SSL}

Representation learning plays an important role in many representation
learning methods are built upon DL, which is a supervised approach. It
is also possible to use an unsupervised approach for representation
learning automatically (i.e. not handcrafted). For example, there exist
correlations between image pixels and their correlations can be removed
using the principal component analysis (PCA).  The application of PCA to
face images was introduced by Turk and Pentland \cite{turk1991face}. The
method is called the ``Eigenface". One main advantage of converting face
images from the spatial domain to the spectral domain is that, when face
images are well aligned, the dimension of input face images can be
reduced significantly and automatically.  Since we attempt to find a
powerful subspace for face image representation, it is a subspace
learning method. 

Chan {\em et al.} \cite{chan2015pcanet} proposed a PCANet that applies
the PCA to input images in two stages. Chen {\em et al.}
\cite{chen2020pixelhop} proposed a PixelHop system that applies cascaded
Saab transforms \cite{kuo2019interpretable} to input images in three
stages, where the Saab transform is a variant of the PCA that adds a
positive bias term to avoid the sign confusion problem
\cite{kuo2016understanding}.  The main difference between Eigenface,
PCANet and PixelHop is to conduct the PCA transform in one, two, or
multiple stages. If we apply one-stage PCA, the face is a pure spatial-
and spectral-domain representations before and after the transform,
respectively.  Since the spatial representation is local, it cannot
offer the global contour and shape information easily.  On the contrary,
the spectral representation is global, it fails to differentiate local
variations. It is desired to get multiple hybrid spatial/spectral
representations. This can be achieved by multi-stage transforms.  Kuo
{\em et al.} developed two multi-stage transforms, called the Saak
transform \cite{kuo2018data} and the Saab transform
\cite{kuo2019interpretable}, respectively.  Recently, the channel-wise
(c/w) Saab transform was proposed in \cite{chen2020pixelhop++} to
enhance the efficiency of the Saab transform. 

\begin{figure*}[t]
\begin{center}
   \includegraphics[width=0.95\linewidth]{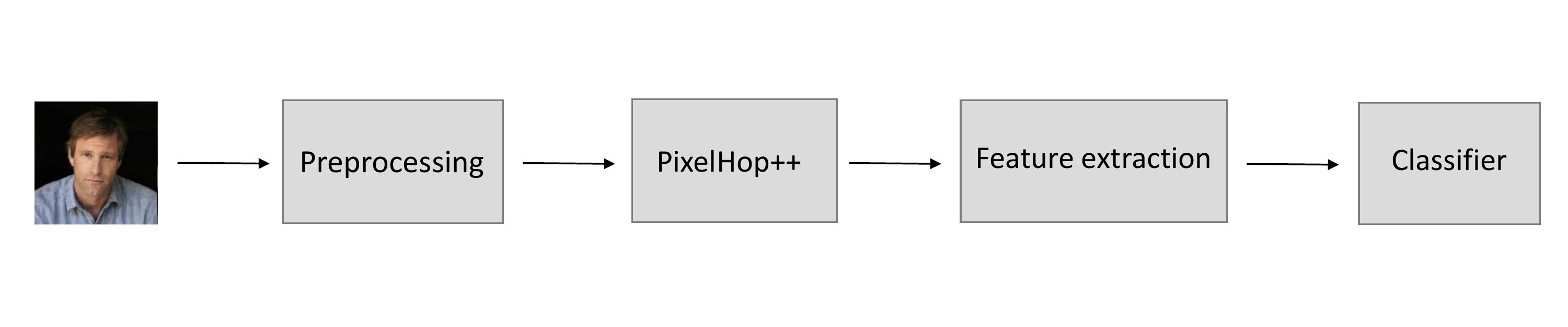}
\end{center}
\caption{An overview of the proposed FaceHop method.}\label{fig:blockdiagram}
\end{figure*}

Inspired by the function of convolutional layers of CNNs
\cite{kuo2019interpretable}, the PixelHop system \cite{chen2020pixelhop}
and the PixelHop++ system \cite{chen2020pixelhop++} were developed to
serve the same function but derived based on a completely different
principle. The weights of convolutional filters in CNNs are obtained by
end-to-end optimization through backpropagation. In contrast, the
convolutional kernels used in PixelHop and PixelHop++ are the Saab
filters. They are derived by exploiting statistical correlations of
neighboring pixels.  As a result, both PixelHop and PixelHop++ are fully
unsupervised.  Neither label nor backpropagation is needed in filter
weights computation. 

The PixelHop++ system \cite{chen2020pixelhop++} is an enhanced version
of the PixelHop system \cite{chen2020pixelhop}.  The main difference
between PixelHop and PixelHop++ is that the former uses the Saab transform
while the latter adopts the c/w Saab transform. The c/w Saab transform
requires fewer model parameters than the Saab transform since channels
are decoupled in the c/w Saab transform. 

\section{Proposed FaceHop Method}\label{sec:method}

An overview of the proposed FaceHop system is shown in Fig.
\ref{fig:blockdiagram}. It consists of four modules: 1) Preprocessing,
2) PixelHop++, 3) Feature extraction, and 4) Classification.  Since
PixelHop++ is the most unique module in our proposed solution for face
gender classification, it is called the FaceHop system.  The
functionality of each module will be explained below in detail. 

\subsection{Preprocessing}\label{subsec:preprocessing} 

Face images have to be well aligned in the preprocessing module to
facilitate their processing in the following pipeline. In this work, we
first use the dlib~\cite{dlib09} tool for facial landmarks localization.
Based on detected landmarks, we apply a proper 2D rotation to each
face image to reduce the effect of pose variation. Then, all face images
are centered and cropped to remove the background. Afterwards, we apply
histogram equalization to each image to reduce the effect of different
illumination conditions.  Finally, all images are resized to a low
resolution one of $32 \times 32$ pixels. 

\subsection{PixelHop++}\label{subsec:preprocessing} 

Both PixelHop and PixelHop++ are used to describe local neighborhoods of
a pixel efficiently and successively.  The size of a neighborhood is
characterized by the hop number.  One-hop neighborhood is the
neighborhood of the smallest size. Its actual size depends on the filter
size. For example, if we use a convolutional filter of size $5 \times
5$, then the hop-1 neighborhood is of size $5 \times 5$.  The Saab
filter weights are obtained by performing dimension reduction on the
neighborhood of a target pixel using PCA. The Saab filters in PixelHop
and PixelHop++ serve as an equivalent role of convolutional filters in
CNNs.  For example, a neighborhood of size $5 \times 5$ has a dimension
of 25 in the spatial domain. We can use the Saab transform to reduce its
original dimension to a significantly lower one. We should mention that
the neighborhood concept is analogous to the receptive field of a
certain layer of CNNs. As we go to deeper layers, the receptive field
becomes larger in CNNs. In the SSL context, we say that the neighborhood
size becomes larger as the hop number increases. 

\begin{figure*}[t]
\begin{center}
\includegraphics[width=.95\linewidth]{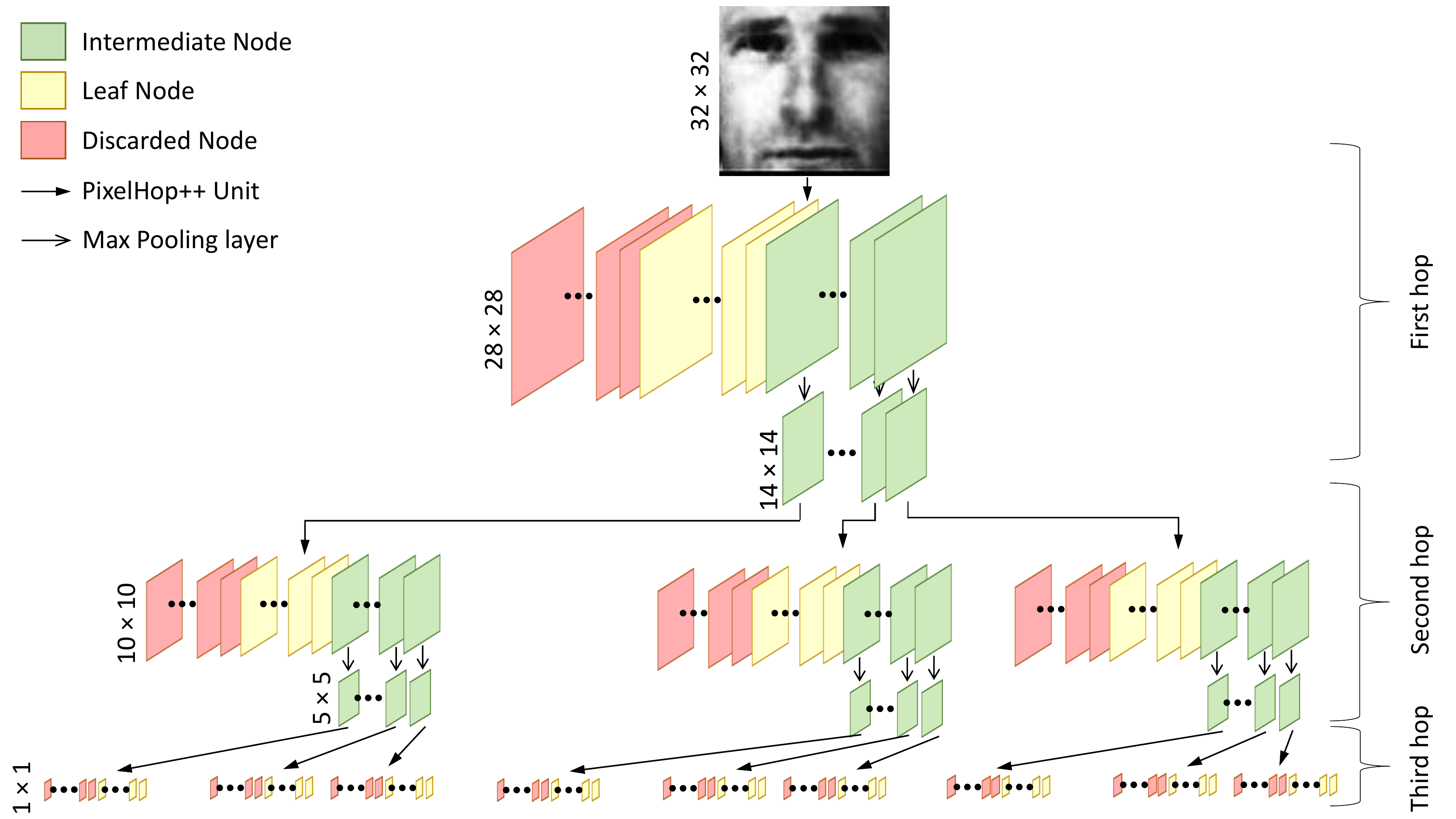}
\end{center}
\caption{Illustration of the proposed 3-hop FaceHop system as a tree-decomposed
representation with its depth equal to three, where each depth layer
corresponds to one hop.}\label{fig:featureextractionmodel}
\end{figure*}

The proposed 3-hop PixelHop++ system is shown in Fig.
\ref{fig:featureextractionmodel}, which is a slight modification of
\cite{chen2020pixelhop++} so as to tailor to our problem.  The input is
a gray-scale face image of size $32\times32$. Each hop consists of a
PixelHop++ unit followed by a $(2\times2)$-to-$(1\times1)$ max-pooling
operation.  A PixelHop++ system has three ingredients: 1) successive
neighborhood construction, 2) channel-wise Saab transform, and 3)
tree-decomposed feature representation. They are elaborated below. 

{\bf 1) Successive neighborhood construction.} We need to specify two
parameters to build the neighborhood of the center pixel at each hop.
There are the window-size and the stride.  We use a window size of
5$\times$5 and stride of 1 in all three hops in Fig.
\ref{fig:featureextractionmodel}. The neighborhood size grows bigger as
the hop number becomes larger due to the max-pooling operation. The
first, second, and third hops characterize the information of the short-,
mid-, and long-range neighborhoods of the center pixel.  Apparently,
each neighborhood has a degree of freedom of 25 in the spatial domain.
By collecting these neighborhood samples from different spatial
locations, we can study their statistical correlations via a covariance
matrix of dimension $25 \times 25$. Then, we conduct the
eigenvector/eigenvalue analysis to the covariance matrix to find a more
economical representation. That is, we can convert pixel values from the
spatial domain to the spectral domain, which leads to the PCA transform,
for dimension reduction. 

{\bf 2) Channel-wise (c/w) Saab transform.} The PCA transform has both
positive and negative responses. We encounter a sign-confusion problem
\cite{kuo2016understanding} when a convolutional operation in the
$(i+1)$th stage has the sum of two terms: 1) a positive response in the
$i$th stage multiplied by a positive outgoing link and 2) a negative
response in the $i$th stage multiplied by a negative outgoing link.
Both terms contribute positive values to the output while their input
patterns are out of phase.  Similarly, there will be another sign-confusion when the convolutional operation in the $(i+1)$th stage has
the sum of a positive response multiplied by a negative filter weight as
well as a negative response multiplied by a positive filter weight. They
both contribute to negative values. 

To resolve such confusion cases, a constant bias term is added to make
all responses positive. This is called the Saab (subspace approximation
via adjusted bias) transform \cite{kuo2019interpretable}.  Typically,
the input of the next pixelhop unit is a 3D tensor of dimension $N_x
\times N_y \times k$, where $N_x=N_y=5$ are spatial dimensions of a
filter and $k$ is the number of kept spectral components.  The Saab
transform is used in PixelHop. 

Since channel responses can be decorrelated by the eigen analysis, we
are able to treat each channel individually.  This results in
channel-wise (c/w) Saab transform \cite{chen2020pixelhop++}.  The main
difference between the standard Saab and the c/w Saab transforms is that
one 3D tensor of dimension $N_x \times N_y \times k$ can be decomposed
into $k$ 2D tensors of dimension $N_x \times N_y$ in the latter.
Furthermore, responses in higher frequency channels are spatially
uncorrelated so that they do not have to go to the next hop.  The c/w
Saab transform is used in PixelHop++. It can reduce the model size
significantly as compared with the Saab transform while preserving the
same performance. 

{\bf 3) Tree-decomposed representation.} Without loss of
generality, we use the first hop to explain the c/w Saab transform
design. The neighborhood of a center pixel contains 25 pixels.  In the
spectral domain, we first decompose it into the direct sum of two
orthogonal subspaces - the DC (direct current) subspace and the AC
(alternating current) subspace. Then, we apply the PCA to the AC
subspace to derive Saab filters. After the first-stage Saab transform,
we obtain one DC coefficient and 24 AC coefficients in a grid of size
$28 \times 28$. We classify AC coefficients into three groups based on
their associated eigenvalues: low-, mid-, and high-frequency AC
coefficients. 

When the eigenvalues are extremely small, we can discard responses in
these channels without affecting the quality of the input face image.
This is similar to the eigenface approach in spirit. For mid-frequency
AC coefficients, the spatial correlation of their responses is too weak
to offer a significant response in hop-2. Thus, we can terminate its
further transform. For low-frequency AC coefficients, the spatial
correlation of their responses is strong enough to offer a significant
response in hop-2. Then, we conduct max-pooling and construct the hop-2
neighborhood of these frequency channels in a grid of size $14 \times 14$.
It is easy to show these hop-by-hop operations using a tree. Then, each
channel corresponds to a node. We use the green, yellow and pink colors
to denote low-, mid- and high-frequency AC channels in Fig.
\ref{fig:featureextractionmodel}. where the DC channel is also colored
in green. They are called the intermediate, leaf, and discard nodes in a
hierarchical tree of depth equal to three. 

To determine which node belongs to which group, we use the energy of
each node as the criterion. The energy of the root node is normalized to
one. The energy of each node in the tree can be computed and normalized
against the energy value of the root node. Then, we can choose two
thresholds (in terms of energy percentages) at each hop to partition
nodes into three types. These energy thresholds are hyperparameters of
the PixelHop++ model. 

\subsection{Feature Extraction}\label{subsec:feature}

Responses at each of the three hops of the FaceHop system have different
characteristics.  As shown in Fig. \ref{fig:featureextractionmodel},
Hop-1 has a response map of size $28 \times 28$, Hop-2 has a response
map of size $10 \times 10$ and Hop-3 has a response map of size $1
\times 1$.  Hop-1 responses give a spatially detailed representation of
the input. Yet, it is difficult for them to offer regional and full
views of the entire face unless the dimension of hop-1 responses becomes
extremely large. This is expensive and unnecessary.  Hop-2 responses
give a coarser view of the entire face so that a small set of them can
cover a larger spatial region. Yet, they do not have face details as
given by Hop-1 responses.  Finally, Hop-3 responses lose all spatial
details but provide a single value at each frequency channel that covers
the full face. The eigenface approach can only capture responses of the
full face and cannot obtain the information offered by hop-1 and hop-2
responses in the FaceHop system. We will extract features based on
responses in all three hops. 

\begin{figure}[t]
\begin{center}
   \includegraphics[width=.9\linewidth]{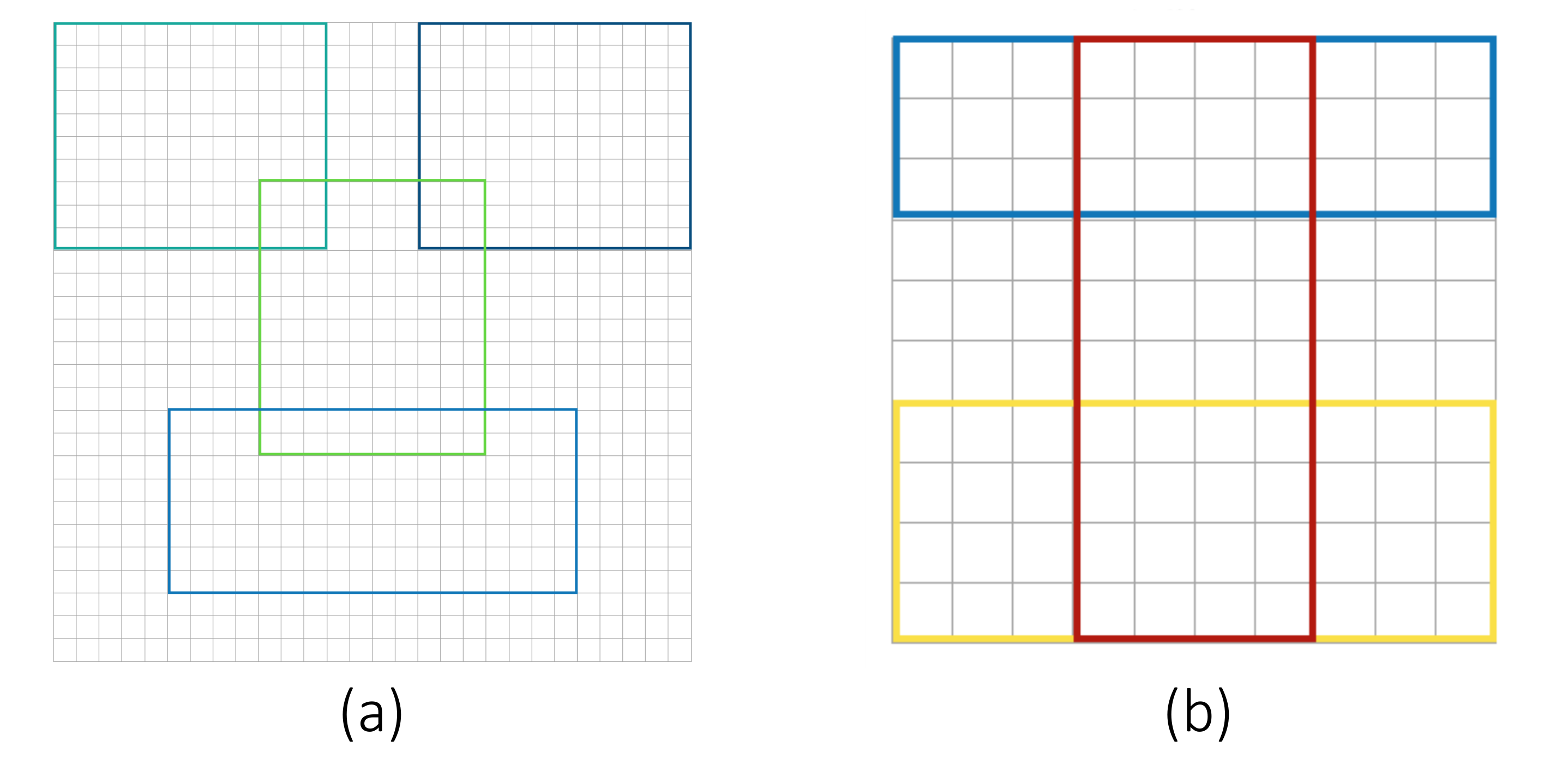}
\end{center}
\caption{Collection of regional responses in hop-1 and hop-2 response
maps as features in the FaceHop system: (a) four regions in hop-1 and
(b) three regions in hop-2.}\label{fig:boundingbox}
\end{figure}

We group pixel responses in hop-1 and hop-2 to form region responses as
shown in Fig. \ref{fig:boundingbox}.
\begin{itemize}

\item {\bf Hop-1.} We collect pixel responses in hop-1 to form four
regions as shown in Fig. \ref{fig:boundingbox} (a). They cover the
left eye, the right eye, the nose, and the mouth regions.  Their spatial
dimensions (height versus width) are $10 \times 12$, $10 \times 12$, $12
\times 10$ and $8 \times 18$, respectively. There are spatial
correlations for responses of the same channel. Thus, we can apply
another PCA to responses of the same hop/region for dimension reduction.
Usually, we can reduce the dimension to the range between 15 and 20.
Afterwards, we concatenate the reduced dimension vector of each region
across all hop-1 channels (including both leaf and intermediate nodes)
to create a hop/region feature vector and feed it to a classifier. There
are four hop-1 regions, and we have four feature vectors that contain
both spatial and spectral information of a face image. The dimension of
hop-1 feature vectors in four regions will be given in Table
\ref{table:feature-dimension}. 
\item {\bf Hop-2.} We collect pixel responses in hop-2 to form three
regions as shown in Fig. \ref{fig:boundingbox} (b). They are: one
horizontal stripe of dimension $3 \times 10$ covering two eyes, another
horizontal stripe of dimension $4 \times 10$ covering the mouth, and one
vertical stripe of dimension $10 \times 4$ covering the nose as well as
the central 40\% region.  Similarly, we can perform dimension reduction
via PCA and concatenate the spatially reduced dimension of each region
across all hop-2 channels to train three classifiers. The dimension of
hop-2 feature vectors in the three regions will be summarized in Table
\ref{table:feature-dimension}. 
\item {\bf Hop-3.} We use all responses of hop-3 as one feature vector
to train a classifier. 
\end{itemize}
It is worthwhile to point out that, although some information of
intermediate nodes will be forwarded to the next hop, different hops
capture different information contents due to varying spatial
resolutions.  For this reason, we include responses in both intermediate
and leaf nodes at hop-1 and hop-2 as features. 

\subsection{Classifiers}\label{subsec:classifier}

As described in Sec. \ref{subsec:feature}, we train four classifiers in
hop 1, another three classifiers in hop 2, and one classifier in hop 3.
Each classifier is a binary classifier. It takes a long feature vector
as the input and makes a soft decision, which is the probability for the
face to be a male or a female. Since the two probabilities add to unity,
we only need to record one of them. Then, at the next stage, we feed
these eight probabilities into a meta classifier for the final decision.
The choice of classifiers can be the Random Forest (RF), the Support
Vector Machine (SVM), and the Logistic Regression (LR). Although the SVM
and the RF classifiers often give higher accuracy, they have a larger
number of model parameters. Since our interest lies in a smaller model
size, we adopt the LR classifier in our experiments only. 

\section{Experiments}\label{sec:experiments}

In this section, we evaluate the proposed FaceHop gender classification
method.  We compare the FaceHop solution with a variant of LeNet-5 in
model sizes and verification performance.  The reason for choosing the
LeNet-5 for performance benchmarking is that it is a small model which is demonstrated to have a relatively high classification accuracy on gray-scale $32 \times 32$ images. The neuron numbers of the modified LeNet-5
model are changed to 16 (1st Conv), 40 (2nd Conv), 140 (1st FC), 60 (2nd
FC) and 2 (output). The modification is needed since human faces are
more complicated than handwritten digits in the MNIST dataset.  For fair comparison, we train both models on the same training data which is achieved by applying the preprocessing and data augmentation to the original face images. We use
only logistic regression (LR) classifiers in FaceHop due to its small
model size.

{\bf Datasets.} We adopt the following two face image datasets in our experiments.
\begin{itemize}
\item \textbf{LFW dataset} \cite{LFWTech} \\
The LFW dataset consists of 13,233 face images of 5,749 individuals,
which were collected from the web. There are 1,680 individuals who have
two or more images.  A 3D aligned version of LFW~\cite{ferrari2016effective} is used in our experiments.
\item \textbf{CMU Multi-PIE dataset} \cite{gross2010multi} \\
The CMU Multi-PIE face dataset contains more than 750,000 images of 337
subjects recorded in four sessions. We select a subset of the 01 session
that contains frontal and slightly non-frontal face images (camera views
05\_0, 05\_1, and 14\_0) with all the available expressions and
illumination conditions in our experiments. 
\end{itemize}

{\bf Data Augmentation.} Since both datasets have significantly fewer
female images, we use two techniques to increase the number of female faces. 
\begin{itemize}
\item Flipping the face images horizontally.
\item Averaging a female face image with its nearest neighbor in the
reduced dimension space to generate a new female face image. To find the
nearest neighbor, we project all female images to a reduced dimension
space, which is obtained by applying PCA and keeping the highest energy
components with 90\% of the total energy.  Dimension reduction is
conducted to eliminate noise and high-frequency components. The quality
of augmented female images is checked to ensure that they are visually
pleasant. 
\end{itemize}
After augmentation, the number of male images is still slightly more than
the number of female images.

{\bf Configuration of PixelHop++.} The configurations of the PixelHop++
module for LFW and CMU Multi-PIE datasets are shown in Table
\ref{table:configuationLFW}. We list the numbers of intermediate nodes, leaf nodes and
discarded nodes at each hop (see Fig. \ref{fig:featureextractionmodel})
in the experiments. In our design, we partition channels into two groups
(instead of three) only at each hop.  That is, they are either discarded
or all forwarded to the next hop.  As a result, there are no leaf nodes
at hop-1 and hop-2. 


\begin{table}[h]
\begin{center}
\begin{tabular}{|c|l|l|l|l|l|l|}
\hline
  &   \multicolumn{3}{|c|}{LFW}  &   \multicolumn{3}{|c|}{CMU Multi-PIE}\\ \cline{2-7}
\multirow{2}{*}{Hop Index}&Interm. & Leaf & Discarded & Interm. & Leaf & Discarded\\ & Node No.& Node No.& Node No.& Node No.& Node No.& Node No.  \\ 
\hline\hline
Hop-1     &       18         &     0         &    7      &       18         &     0         &    7          \\ \hline
Hop-2     &      122         &     0         &   328   &      117         &     0         &   333           \\ \hline
Hop-3     &        0         &    233        &  2,817   &   0         &    186        &  2,739          \\ \hline
\end{tabular}
\end{center}
\caption{Configurations of PixelHop++ for LFW and CMU Multi-PIE..}
\label{table:configuationLFW}
\end{table}

{\bf Feature Vector Dimensions of Varying Hop/Region Combinations.} The
dimensions of feature vectors of varying hop/region combinations are
summarized in Table \ref{table:feature-dimension}. As discussed earlier,
hop-1 has 4 spatial regions, hop-2 has three spatial regions and all
nodes of hop-3 form one feature vector. Thus, there are eight
hop/region combinations in total.  Since there are spatial correlations in
regions given in Fig.  \ref{fig:boundingbox}, we apply PCA to regional
responses collected from all channels and keep leading components for
dimension reduction. We keep 15 components for the LFW dataset and 20
components for the CMU Multi-PIE datasets, respectively. Then, the
dimension of each feature vector at hop-1 and hop-2 is the product of 15
(or 20) and the sum of intermediate and leaf nodes at the associated hop
for the LFW (or CMU Multi-PIE) dataset. 

\begin{table}[h]
\begin{center}
{\scriptsize
\begin{tabular}{|c|c|c|c|c|c|}\hline
Hop/Region           & LFW & MPIE     & Hop/Region             & LFW   & MPIE   \\ \hline\hline
Hop-1 (left eye)     & 270 & 360      & Hop-2 (upper stripe)   & 1,830 & 2,340  \\ \hline
Hop-1 (right eye)    & 270 & 360      & Hop-2 (lower stripe)   & 1,830 & 2,340  \\ \hline
Hop-1 (nose)         & 270 & 360      & Hop-2 (vertical strip) & 1,830 & 2,340  \\ \hline
Hop-1 (mouth)        & 270 & 360      & Hop-3                  & 233   & 186    \\ \hline
\end{tabular}}
\end{center}
\caption{Feature vector dimensions for LFW and CMU Multi-PIE.} 
\label{table:feature-dimension}
\end{table}

{\bf Performance and Model Size Comparison for LFW.} We randomly partition male
and original plus augmented female images in the LFW dataset into 80\%
(for training) and 20\% (for testing) two sets individually. Then, they
are mixed again to form the desired training and testing datasets. This
is done to ensure the same gender percentages in training and testing.
We train eight individual hop/region LR classifiers and one meta LR
classifier for ensembles. Then, we apply them to the test data to find
out their performance.  We repeat the same process four times to get the
mean testing accuracy and the standard deviation value, and report the
testing performance of each individual hop/region in Table
\ref{table:acc-LFW-individual}. 

\begin{table}[h]
\begin{center}
{\scriptsize
\begin{tabular}{|c|c|c|c|}\hline
Classifier           & Accuracy$(\%)$   & Classifier & Accuracy $(\%)$ \\ \hline\hline
Hop-1 (left eye)     & 86.70 $\pm$ 0.65   & Hop-2 (upper stripe)   & 92.25 $\pm$ 0.22 \\ \hline
Hop-1 (right eye)    & 86.14 $\pm$ 0.66   & Hop-2 (lower stripe)   & 89.70 $\pm$ 0.73 \\ \hline
Hop-1 (nose)         & 82.90 $\pm$ 0.61   & Hop-2 (vertical strip) & 92.42 $\pm$ 0.56 \\ \hline
Hop-1 (mouth)        & 83.42 $\pm$ 0.74   & Hop-3                  & 91.22 $\pm$ 0.46 \\ \hline
\end{tabular}}
\end{center}
\caption{Performance comparison of each individual hop/region classifier 
for LFW.}\label{table:acc-LFW-individual}
\end{table}

The mean testing accuracy ranges from 82.90\% (hop-1/nose) to 92.42\%
(hop-2/vertical stripe). The standard deviation is relatively small.
Furthermore, we see that hop-2 and hop-3 classifiers perform better than
hop-1 classifiers. Based on this observation, we consider two ensemble
methods. In the first scheme, called FaceHop I, we fuse soft decisions
of all eight hop/region classifiers with a meta classifier.  In the
second scheme, called FaceHop II, we only fuse soft decisions of four
hop/region classifiers from hop-2 and hop-3 only. The testing accuracy
and the model sizes of LeNet-5, FaceHop I and FaceHop II are compared in
Table \ref{table:acc-LFW}. FaceHop I and FaceHop II outperform LeNet-5
in terms of classification accuracy by 0.70\% and 0.86\%, respectively,
where their model sizes are only about 33.7\% and 22.2\% of LeNet-5.
Clearly, FaceHop II is the favored choice among the three for its
highest testing accuracy and smallest model size. 

\begin{table}[h]
\begin{center}
\begin{tabular}{|l|c|c|}\hline
Method & Accuracy $(\%)$ & Model Size \\ \hline\hline
LeNet-5                           & 93.77 $\pm$ 0.43 & 75,846 \\ \hline
FaceHop I (all three hops)        & 94.47 $\pm$ 0.54 & 25,543 \\ \hline
FaceHop II (hop-2 \& hop-3 only)  & 94.63 $\pm$ 0.47 & 16,895 \\ \hline
\end{tabular}
\end{center}
\caption{Performance comparison of LeNet-5, FaceHop I and FaceHop II in
accuracy rates and model sizes for LFW.}\label{table:acc-LFW}
\end{table}

{\bf Performance and Model Size Comparison for CMU Multi-PIE.} Next, we
show the classification accuracy of each individual hop/region classifier
for the CMU Multi-PIE dataset in Table \ref{table:acc-CMU-individual}.
Their accuracy values range from 63.02\% (hop-1/mouth) to 91.95\%
(hop-2/upper stripe). It appears that CMU Multi-PIE is more challenging
than LFW if we focus on the performance of each individual classifier by
comparing Tables \ref{table:acc-LFW-individual} and
\ref{table:acc-CMU-individual}. 

\begin{table}[htb]
\begin{center}
{\scriptsize
\begin{tabular}{|c|c|c|c|}\hline
Classifier           & Accuracy $(\%)$  & Classifier             & Accuracy $(\%)$ \\ \hline\hline
Hop-1 (left eye)     &   79.33$\pm$0.33   & Hop-2 (upper stripe)   & 91.95$\pm$0.18    \\ \hline
Hop-1 (right eye)    &   78.64$\pm$0.25   & Hop-2 (lower stripe)   & 87.00$\pm$0.15    \\ \hline
Hop-1 (nose)         &   65.19$\pm$0.36   & Hop-2 (vertical strip) & 91.34$\pm$0.22    \\ \hline
Hop-1 (mouth)        &   63.02$\pm$0.41   & Hop-3                  & 84.55$\pm$0.77    \\ \hline
\end{tabular}}
\end{center}
\caption{Performance comparison of each individual hop/region classifier
for CMU Multi-PIE.} \label{table:acc-CMU-individual}
\end{table}

We consider two ensemble schemes as done before. FaceHop I uses all
eight soft decisions while FaceHop II takes only four soft decisions
from hop-2 and hop-3.  The mean accuracy performance of LeNet-5, FaceHop
I and FaceHop II are compared in Table \ref{table:acc-CMU}. It is
interesting to see that FaceHop I and II have slightly better ensemble
results of CMU Multi-PIE than of LFW, respectively. The performance of
LeNet-5 also increases from 93.77\% (LFW) to 95.08\% (CMU Multi-PIE). As
far as the model size is concerned, the model sizes of FaceHop I and
FaceHop II are about 38.4\% and 23.2\% of LeNet-5, respectively.  Again,
FaceHop II is the most favored solution among the three for its highest
testing accuracy and smallest model size. 
\begin{table}[htb]
\begin{center}
\begin{tabular}{|l|c|c|}\hline
Method                            & Accuracy $(\%)$     & Model Size \\ \hline\hline
LeNet-5                           & 95.08              & 75,846 \\ \hline
FaceHop I (all three hops)        & 95.09 $\pm$ 0.24    & 29,156 \\ \hline
FaceHop II (hop-2 and hop-3 only) & 95.12 $\pm$ 0.26    & 17,628 \\ \hline
\end{tabular}
\end{center}
\caption{Performance comparison of LeNet-5, FaceHop I and FaceHop II in
accuracy rates and model sizes for CMU Multi-PIE.}\label{table:acc-CMU}
\end{table}
\section{Conclusion and Future Work}\label{sec:conclusion}

A light-weight low-resolution face gender classification method, called
FaceHop, was proposed. This solution finds applications in
resource-constrained environments with limited networking and computing.
FaceHop has
several desired characteristics, including a small model size, a small
training data amount, low training complexity, and low-resolution input
images. The effectiveness of the FaceHop method for
gender classification was demonstrated by experiments on two
benchmarking datasets. 
 
In this paper, we demonstrated the potential of the SSL principle for effective feature extraction from face images. As to future work, we would like to test more datasets for gender classification and also extend the SSL principle for identifying heterogeneous and correlated face
attributes such as gender, age, and race. It is particularly interesting
to develop a multi-task learning approach. Furthermore, it will be
desired to work on high-resolution face images and see whether we can
get significant performance improvement using the SSL principle in
classification accuracy, computational complexity, and memory usage. 
%
%
%
 \bibliographystyle{IEEEtran}
 \bibliography{samplepaper}

\end{document}